\documentclass[11pt,a4paper]{article}
\usepackage[hyperref]{naaclhlt2018}
\usepackage{times}
\usepackage{latexsym}
\usepackage{graphicx}
\usepackage{url}
\usepackage{adjustbox}
\usepackage{amssymb}
\usepackage{amsmath}

\aclfinalcopy 

\DeclareGraphicsExtensions{.pdf,.png,.jpg}

\title{Stochastic Answer Networks for SQuAD 2.0}

\author{
Xiaodong Liu$^\bold{\dagger}$,
Wei Li$^\bold{\dagger}$,
Yuwei Fang$^\bold{\dagger}$,
Aerin Kim$^\bold{\dagger}$,
Kevin Duh$^\ddagger$
and Jianfeng Gao$^\bold{\dagger}$ \\
  $^\bold{\dagger}$    
  Microsoft Research, Redmond, WA, USA \\
   $^\bold{\ddagger}$
  Johns Hopkins University, Baltimore, MD, USA \\
 {\tt $^\bold{\dagger}$\{xiaodl,wli,yuwfan,ahkim,jfgao\}@microsoft.com
    $^\bold{\ddagger}$kevinduh@cs.jhu.edu}
}
\date{}

\begin{document}
\maketitle
\begin{abstract}
This paper presents an extension of the Stochastic Answer Network (SAN), one of the state-of-the-art machine reading comprehension models, to be  able to judge whether a question is unanswerable or not. 
The extended SAN contains two components: a span detector and a binary classifier for judging whether the question is unanswerable, and both components are jointly optimized.
Experiments show that SAN achieves the results competitive to the state-of-the-art on Stanford Question Answering Dataset (SQuAD) 2.0. To facilitate the research on this field, we release our code: \href{https://github.com/kevinduh/san\_mrc}{https://github.com/kevinduh/san\_mrc}.
\end{abstract}

\section{Background}
\label{sec:bck}
Teaching machine to read and comprehend a given passage/paragraph and answer its corresponding questions is a challenging task. It is also one of the long-term goals of natural language understanding, and has important applications in e.g., building intelligent agents for conversation and customer service support.  In a real world setting, it is necessary to judge whether the given questions are answerable given the available knowledge, and then generate correct answers for the ones which are able to infer an answer in the passage or an empty answer (as an unanswerable question) otherwise.
\begin{figure}[t]
    \centering
\adjustbox{trim={.05\width} {.01\height} {.01\width} {.01\height},clip}{\includegraphics[scale=0.7]{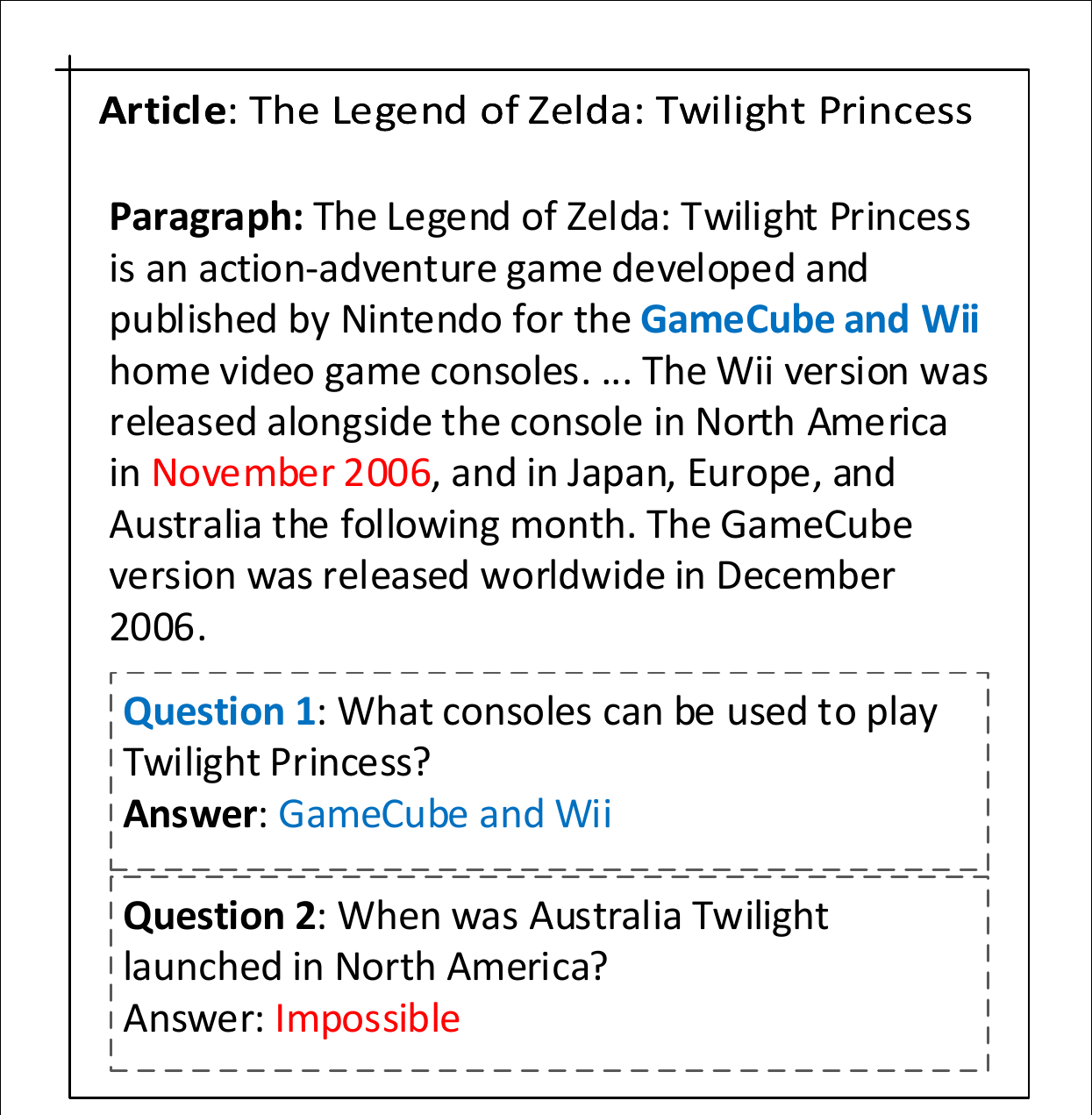}}
\caption{\label{fig:sample} Examples from SQuAD v2.0. The first question is answerable which indicates its answer highlighted in blue can be found in the paragraph; while the second question is unanswerable and its \textit{plausible} answer is highlighted in red.
} 
\end{figure} 

In comparison with many existing MRC systems \cite{match-lstm2016, san2018, qanet2018,bidaf2016, reasonet++2017}, which extract answers by finding a sub-string in the passages/paragraphs, we propose a model that not only extracts answers but also predicts whether such an answer should exist.
Using a multi-task learning approach (c.f.~\cite{mlt2015}), we extend the Stochastic Answer Network (SAN) \cite{san2018} for MRC answer span detector to include a classifier that whether the question is unanswerable.
The unanswerable classifier is a pair-wise classification model \cite{liu2018stochastic} which predicts a label indicating whether the given pair of a passage and a question is unanswerable. The two models share the same lower layer to save the number of parameters, and separate the top layers for different tasks (the span detector and binary classifier).
\begin{figure*}[t!]
\centering
\adjustbox{trim={.01\width} {.02\height} {.05\width} {.02\height},clip}%
  {\includegraphics[scale=0.78]{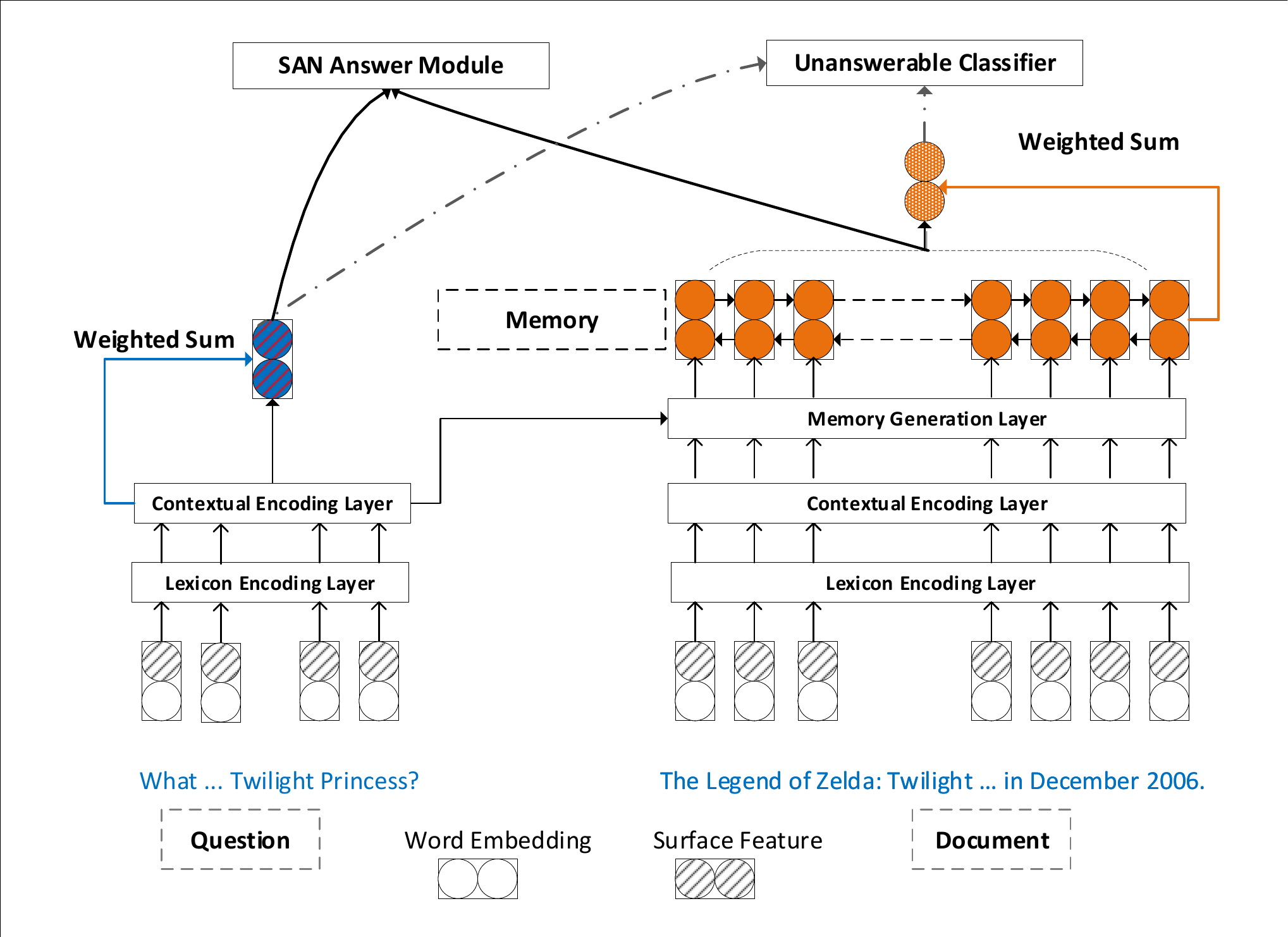}}
\caption{\label{fig:model} {\bf Architecture of the proposed model for Reading Comprehension:} It includes two components: a span detector (the upper left SAN answer module) and an unanswerable classifier (the upper right module). It contains two sets of layers: the shared layers including a lexicon encoding layer, contextual encoding layer and memory generation layer; and the task specific layers including the SAN answer module for span detection, and a binary classifier determining whether the question is unanswerable. The model is learned jointly.}
\end{figure*}
Our model is pretty simple and intuitive, yet efficient. Without relying on the large pre-trained language models (ELMo) \cite{elmo2018}, the proposed model achieves competitive results to the state-of-the-art on Stanford Question Answering Dataset (SQuAD) 2.0. 

The contribution of this work is summarized as follows. First, we propose a simple yet efficient model for MRC that handles unanswerable questions and is  optimized jointly. Second, our model achieves competitive results on SQuAD v2.0. 

\section{Model}
\label{sec:model}
The Machine Reading Comprehension is a task which takes a question $Q=\{q_0, q_1, ..., q_{m-1}\}$ and a passage/paragraph $P=\{p_0, p_1, ..., p_{n-1}\}$ as inputs, and aims to find an answer span $A$ in $P$. We assume that if the question is answerable, the answer $A$ exists in $P$ as a contiguous text string; otherwise, $A$ is an empty string indicating an unanswerable question. Note that to handle the unanswerable questions, we manually append a dumpy text string \textit{NULL} at the end of each corresponding passage/paragraph. Formally, the answer is formulated as $A=\{a_{begin}, a_{end}\}$. In case of unanswerable questions, $A$ points to the last token of the passage.

Our model is a variation of SAN \cite{san2018}, as shown in Figure~\ref{fig:model}. The main difference is the additional binary classifier added in the model justifying whether the question is unanswerable. Roughly, the model includes two different layers: the shared layer and task specific layer. The shared layer is almost identical to the lower layers of SAN, which has a lexicon encoding layer, a contextual layer and a memory generation layer. On top of it, there are different answer modules for different tasks. We employ the SAN answer module for the span detector and a one-layer feed forward neural network for the binary classification task. It can also be viewed as a multi-task learning \cite{multitask1997,mlt2015,multimrc2018}. We will briefly describe the model from ground up as follows. Detailed descriptions can be found in \cite{san2018}.

\noindent \textbf{Lexicon Encoding Layer.}  We map the symbolic/surface feature of $P$ and $Q$ into neural space via word embeddings \footnote{We use 300-dim GloVe \cite{glove2014} vectors.}, 16-dim part-of-speech (POS) tagging embeddings, 8-dim named-entity embeddings and 4-dim hard-rule features\footnote{It includes 3 matching features which are determined based on the original word, lower case, and lemma, respectively, and one term sequence feature.}. Note that we use small embedding size of POS and NER to reduce model size and they mainly serve the role of coarse-grained word clusters. Additionally, we use question enhanced passages word embeddings which can viewwed as soft matching between questions and passages. At last, we use two separate two-layer position-wise Feed-Forward Networks (FFN) \cite{vaswani2017attention, san2018} to map both question and passage encodings into the same dimension. As results, we obtain the final lexicon embeddings for the tokens for $Q$ as a matrix $E^q\in  \mathbb{R}^{d \times m}$, and tokens in $P$ as $E^q\in  \mathbb{R}^{d \times n}$.

\noindent \textbf{Contextual Encoding Layer.} A shared two-layers BiLSTM is used on the top to encode the contextual information of both passages and questions. To avoid overfitting, we concatenate a pre-trained 600-dimensional CoVe vectors\footnote{https://github.com/salesforce/cove} \cite{mccann2017learned} trained on German-English machine translation dataset, with the aforementioned lexicon embeddings as the final input of the contextual encoding layer, and also with the output of the first contextual encoding layer as the input of its second encoding layer. Thus, we obtain the final representation of the contextual encoding layer by a concatenation of the outputs of two BiLSTM: $H^q\in \mathbb{R}^{4d\times m}$ for questions and $H^p\in \mathbb{R}^{4d\times n}$ for passages.

\noindent \textbf{Memory Generation Layer.} In this layer, we generate a working memory by fusing information from both passages $H^p$ and questions $H^q$. The attention function \cite{vaswani2017attention} is used to compute the similarity score between passages and questions as:
\[C=\text{dropout}\left(f_{\text{attention}}(\hat{H}^q, \hat{H}^p)\right) \in\mathbb{R}^{m\times n}.  \]
Note that $\hat{H^q}$ and $\hat{H^p}$ is transformed from $H^q$ and $H^p$ by one layer neural network $ReLU(Wx)$, respectively. 
A question-aware passage representation is computed as
$U^p = \text{concat}(H^p, H^qC)$. After that, we use the method of \cite{selfatt2017} to apply self attention to the passage:
\[\hat{U}^p = U^p\text{drop}_{\text{diag}}(f_{\text{attention}}(U^p, U^p)), \]
where $\text{drop}_{\text{diag}}$ means that we only drop diagonal elements on the similarity matrix (i.e., attention with itself). At last, $U^p$ and $\hat{U}^p$ are concatenated and are passed through a BiLSTM to form the final memory: $M=\text{BiLSTM}([U^p];\hat{U}^p])$.

\noindent \textbf{Span detector.} We adopt a multi-turn answer module for the span detector \cite{san2018}. Formally, at time step $t$ in the range of $\{1, 2, ..., T-1\}$, the state is defined by $s_t = GRU(s_{t-1}, x_t)$. The initial state $s_0$ is the summary of the $Q$: $s_0=\sum_j \alpha_j H^q_{j}$, where $\alpha_j = \frac{exp(w_0 \cdot H^q_j)}{\sum_{j'}exp(w_0 \cdot H^q_{j'})}$.
Here, $x_t$ is computed from the previous state $s_{t-1}$ and memory $M$: $x_t=\sum_j\beta_j M_j$ and $\beta_j = softmax(s_{t-1}W_1M)$. 
Finally, a bilinear function is used to find the begin and end point of answer spans at each reasoning step $t \in \{0,1,\ldots,T-1\}$:
\begin{equation}
P_t^{begin} = softmax(s_tW_2M)
\label{eq:begin}
\end{equation} 
\begin{equation}
P_t^{end} = softmax(s_tW_3M)\footnote{Note that we use a simper formula in Eq~\ref{eq:end} as \cite{san2018}.}.
\label{eq:end}
\end{equation}
The final prediction is the average of each time step: $P^{\text{begin}}=\frac{1}{T}\sum_{t}P_t^{\text{begin}},P^{\text{end}}=\frac{1}{T}\sum_{t}P_t^{\text{end}}$. We randomly apply dropout on the step level in each time step during training, as done in \cite{san2018}.

\noindent \textbf{Unanswerable classifier.} We adopt a one-layer neural network as our unanswerable binary classifier: \begin{equation}
P^{u} = sigmoid([s_0;m_0]W_4)
\label{eq:class}
\end{equation}
, where $m_0$ is the summary of the memory: $m_0=\sum_j \gamma_j M_{j}$, where $\gamma_j = \frac{exp(w_5 \cdot M_j)}{\sum_{j'}exp(w_5 \cdot M_{j'})}$. $P^u$ denotes the probability of the question which is unanswerable.  

\noindent \textbf{Objective}
The objective function of the joint model has two parts: 
\begin{equation}
\mathcal{L}_{joint} = \mathcal{L}_{span} + \lambda \mathcal{L}_{classifier}
\label{eq:joint}
\end{equation}
Following \cite{match-lstm2016}, the span loss function is defined:
\begin{equation}
\mathcal{L}_{span} = -(\log P^{begin} + \log P^{end}).
\label{eq:lspan}
\end{equation}
The objective function of the binary classifier is defined:
\begin{equation}
\mathcal{L}_{classifier} = -\{y \ln P^u + (1-y) \ln(1-P^u)\}
\label{eq:lc}
\end{equation}
where $y \in \{0,1\}$ is a binary variable: $y=1$ indicates the question is unanswerable and $y=0$ denotes the question is answerable.

\section{Experiment}
\label{sec:exp}
\subsection{Setup}
We evaluate our system on SQuAD 2.0 dataset \cite{squadv22018}, a new MRC dataset which is a combination of Stanford Question Answering Dataset (SQuAD) 1.0 \cite{squad2016} and additional unanswerable question-answer pairs.
The answerable pairs are around 100K; while the unanswerable questions are around 53K. This dataset contains about 23K passages and they come from approximately 500 Wikipedia articles. All the questions and answers are obtained by crowd-sourcing. 
Two evaluation metrics are used: Exact Match (\textbf{EM}) and Macro-averaged F1 score (\textbf{F1}) \cite{squadv22018}.
\subsection{Implementation details}
\label{sec:imp}
We utilize spaCy\footnote{\href{https://spacy.io}{https://spacy.io}} tool to tokenize the both passages and questions, and generate lemma, part-of-speech and named entity tags. The word embeddings are initialized with pre-trained 300-dimensional GloVe \cite{glove2014}. A 2-layer BiLSTM is used encoding the contextual information of both questions and passages. Regarding the hidden size of our model, we search greedily among $\{128, 256, 300\}$.  During training, Adamax \cite{kingma2014adam} is used as our optimizer. The min-batch size is set to 32. The learning rate is initialized to 0.002 and it is halved after every 10 epochs. The dropout rate is set to 0.1. To prevent overfitting, we also randomly set 0.5\% words in both passages and questions as unknown words during the training. Here, we use a special token \textit{unk} to indicate a word which doesn't appear in GloVe.  $\lambda$ in Eq~\ref{eq:joint} is set to 1.

\section{Results}
\label{sec:res}
We would like to investigate effectiveness the proposed joint model. To do so, the same shared layer/architecture is employed in the following variants of the proposed model:
\begin{enumerate}
    \item SAN: it is standard SAN model \footnote{To handle the unanswerable questions, we append a \textit{NULL} string at the end of the passages for the unanswerable questions.} \cite{san2018}, which is trained by using Eq~\ref{eq:lspan}.
    \item Joint SAN: the proposed joint model Eq~\ref{eq:joint}.
    \item Joint SAN + Classifier: the proposed joint model Eq~\ref{eq:joint} which also use the output information from the unanswerable binary classifier \footnote{We set the answer to an empty string if the output probability of the classifier is larger than 0.5.}.
\end{enumerate}
\begin{table}[t]
\centering
\begin{tabular}{l|c| c}
\hline \hline
Single model:&EM& F1 \\ \hline
\hline
SAN & 67.89& 70.68\\ \hline
Joint SAN &69.27 & 72.20 \\ \hline
Joint SAN + Classifier&\textbf{69.54} & \textbf{72.66} \\ \hline \hline
\end{tabular}
\caption{\label{tab:joint} Performance on the SQuAD 2.0 development dataset.}
\end{table}
The results in terms of EM and F1 is summarized in Table ~\ref{tab:joint}. We observe that Joint SAN outperforms the SAN baseline with a large margin, e.g., 67.89 vs 69.27 (+1.38) and 70.68 vs 72.20 (+1.52) in terms of EM and F1 scores respectively, so it demonstrates the effectiveness of the joint optimization. By incorporating the output information of classifier into Joint SAN, it obtains a slight improvement, e.g., 72.2 vs 72.66 (+0.46) in terms of F1 score. By analyzing the results, we found that in most cases when our model extract an \textit{NULL} string answer, the classifier also predicts it as an unanswerable question with a high probability.

\begin{table}[ht!]
\centering
\begin{tabular}{l|c| c}
\hline \hline
\multicolumn{3}{c}{SQuAD 2.0 \textbf{development} dataset} \\ \hline
&EM& F1 \\ \hline
BNA$^1$& 59.8 & 62.6 \\ \hline
DocQA$^1$& 61.9 & 64.8 \\ \hline
R.M-Reader$^2$& 66.9 & 69.1 \\ \hline
R.M-Reader + Verifier$^2$& 68.5 & 71.5 \\ \hline
Joint SAN&\textbf{69.3} & \textbf{72.2} \\ \hline \hline
\multicolumn{3}{c}{SQuAD 2.0 \textbf{development} dataset + \textbf{ELMo}} \\ \hline
DocQA$^1$& 65.1 & 67.6 \\ \hline
R.M-Reader + Verifier$^2$& \textbf{72.3} & \textbf{74.8} \\ \hline
\multicolumn{3}{c}{SQuAD 2.0 \textbf{test} dataset} \\ \hline \hline
BNA$^1$& 59.2 & 62.1 \\ \hline
DocQA$^1$& 59.3 & 62.3 \\ \hline
DocQA + ELMo$^1$& 63.4 & 66.3 \\ \hline
 R.M-Reader$^{2*}$ & \textbf{71.7}&\textbf{74.2}\\ \hline
Joint SAN$^{\#}$&68.7 & 71.4 \\ \hline \hline
\end{tabular}
\caption{\label{tab:leaderboard} Comparison with published results in literature. $^1$: results are extracted from \cite{squadv22018}; $^2$: results are extracted from \cite{verify2018}. $^*$: it is unclear which model is used. $^\#$: we only evaluate the Joint SAN in the submission.}
\end{table}
Table~\ref{tab:leaderboard} reports comparison results in literature published \footnote{For the full leaderboard results, please refer to \href{https://rajpurkar.github.io/SQuAD-explorer.}{https://rajpurkar.github.io/SQuAD-explorer}}.
Our model achieves state-of-the-art on development dataset in setting without pre-trained large language model (ELMo). Comparing with the much complicated model R.M.-Reader + Verifier, which includes several components, our model still outperforms by 0.7 in terms of F1 score. Furthermore, we observe that ELMo gives a great boosting on the performance, e.g., 2.8 points in terms of F1 for DocQA. This encourages us to incorporate ELMo into our model in future.

\noindent \textbf{Analysis.} To better understand our model, we analyze the accuracy of the classifier in our joint model. We obtain 75.3 classification accuracy on the development with the threshold 0.5. By increasing value of $\lambda$ in Eq~\ref{eq:joint}, the classification accuracy reached to 76.8 ($\lambda=1.5$), however the final results of our model only have a small improvement (+0.2 in terms of F1 score). It shows that it is important to make balance between these two components: the span detector and unanswerable classifier.
\section{Conclusion}
To sum up, we proposed a simple yet efficient model based on SAN. It showed that the joint learning algorithm boosted the performance on SQuAD 2.0. We also would like to incorporate ELMo into our model in future.
\section*{Acknowledgments}
We thank Yichong Xu, Shuohang Wang and Sheng Zhang for valuable discussions
and comments. We also thank Robin Jia for the help on SQuAD evaluations.
\bibliography{mrc}
\bibliographystyle{acl_natbib}

\end{document}